\documentclass{article}
\usepackage{spconf,amsmath,epsfig}
\usepackage[utf8]{inputenc} 
\usepackage[T1]{fontenc}    
\usepackage{hyperref}       
\usepackage{url}            
\usepackage{booktabs}       
\usepackage{amsfonts}       
\usepackage{nicefrac}       
\usepackage{microtype}      
\usepackage{multirow}
\usepackage{graphicx}
\usepackage{color}
\usepackage{subfigure}
\usepackage{cleveref}
\usepackage{fancyhdr}

\let\OLDthebibliography\thebibliography
\renewcommand\thebibliography[1]{
  \OLDthebibliography{#1}
  \setlength{\parskip}{0pt}
  \setlength{\itemsep}{0pt plus 0.3ex}
}
\fancypagestyle{copyright}{%
    \fancyhf{}
    \cfoot{\footnotesize%
        \textcopyright~2022 IEEE. Personal use of this
        material is permitted. Permission from IEEE must be obtained for all other
        uses, in any current or future media, including reprinting\slash republishing
        this material for advertising or promotional purposes, creating new collective
        works, for resale or redistribution to servers or lists, or reuse of any
        copyrighted component of this work in other works. This material is referenced
        by DOI: {\href{https://doi.org/XXXXXXXXXXXXX}{XXXXXXXXXXXXXXXXXX}}
    }
}

\title{Monitoring Urban Forests from Auto-Generated Segmentation Maps}
\name{%
Conrad M Albrecht$^{1,2}$,
Chenying Liu$^{1,2}$,
Yi Wang$^{1,2}$,
Levente Klein$^3$,
Xiao Xiang Zhu$^{1,2}$\thanks{This work is supported by Helmholtz Association’s Initiative and Networking Fund through Helmholtz AI.}
}
\address{%
$^1$Remote Sensing Technology Institute, German Aerospace Center, Germany\\
$^2$Data Science in Earth Observation, Technical University of Munich, Germany\\
$^3$TJ Watson Research Center, IBM Research, Yorktown Heights, NY, USA
}
\begin{document}
%
\maketitle
\begin{abstract}
  We present and evaluate a weakly-supervised methodology to quantify the
  spatio-temporal distribution of urban forests based on remotely
  sensed data with close-to-zero human interaction. Successfully
  training machine learning models for semantic segmentation typically depends
  on the availability of high--quality labels. We evaluate the benefit of
  high--resolution, three-dimensional point cloud data (LiDAR) as source of
  noisy labels in order to train models for the localization of trees in orthophotos.
  As proof of concept we sense Hurricane \textit{Sandy}'s impact on urban
  forests in Coney Island, New York City (NYC) and reference it to less impacted urban
  space in Brooklyn, NYC.
\end{abstract}
\thispagestyle{copyright}
%
\section{Introduction \& Motivation}

While overhead imagery is available in many parts of the world \cite{morelli2016managing},
the lack of associated labels makes image classification and semantic segmentation challenging.
To manually annotate historical datasets at the granularity of a single tree, individual buildings, secondary roads,
etc., exceeds the capacity of the scientific community. Often, patterns and shapes can be easily
recognized by eye in high-resolution orthophotos. However, developing automatic labeling
tools with similar performance is a challenge. In \cite{autogeolabel} efforts focus on automatically
extracting features from the spatial statistics of remote sensing measurements. Based on a
comprehensive set of humanly-defined rules, massive amounts of noisy segmentation maps
automatically get generated as basis for e.g.\ weakly--supervised machine learning models
to label high-resolution aerial imagery.

While automatic labels may bear noise, such information can prove adequate for semantic
segmentation in order to track change of land use, or to quantify the impact of
natural disasters and climate change. 
Noisy labels in binary classification \cite{natarajan2013learning}, image classification
\cite{xiao2015learning}, and deep learning \cite{han2019deep} did get studied extensively.
Similar approaches in the geospatial domain \cite{wang2020weakly,schmitt2020weakly} 
are less accurate and require further refinements  \cite{robinson2021global,dong2021high}.

Many urban areas are located in coastal regions; exposing them to flooding,
hurricane damage, and surging waves vulnerabilities. In a case study we explore the automated,
rule-based generation of noisy labels for urban forests from high--density, three-dimensional
point cloud data. Subsequently, we employ those auto-generated labels to train a convolutional
neural network (U-Net, \cite{ronneberger2015u}) and a standard, noise-robust machine learning model
(Support Vector Machine, SVM, \cite{cortes1995support}) to semantically segment historical
multi-spectral orthophotos for tree identification.

In summary, our work demonstrates the value of generating \textit{noisy}, but \textit{high--density}
labels for remote sensing machine learning models to accurately identify urban forests in historical imagery
with little human interaction. In particular, we extract noisy labels from derived statistics
of a \textit{Light Detection And Ranging} (LiDAR) survey in the area of New York City. Then,
we utilize these labels for semantic segmentation of multi-spectral imagery to identify damage
from Hurricane \textit{Sandy}.

\section{Datasets}
\label{sec:datasets}
\begin{figure*}[t!]
  \centering
  \subfigure[orthophoto]{\label{subfig:orthophoto}
      \includegraphics[width=.23\textwidth]{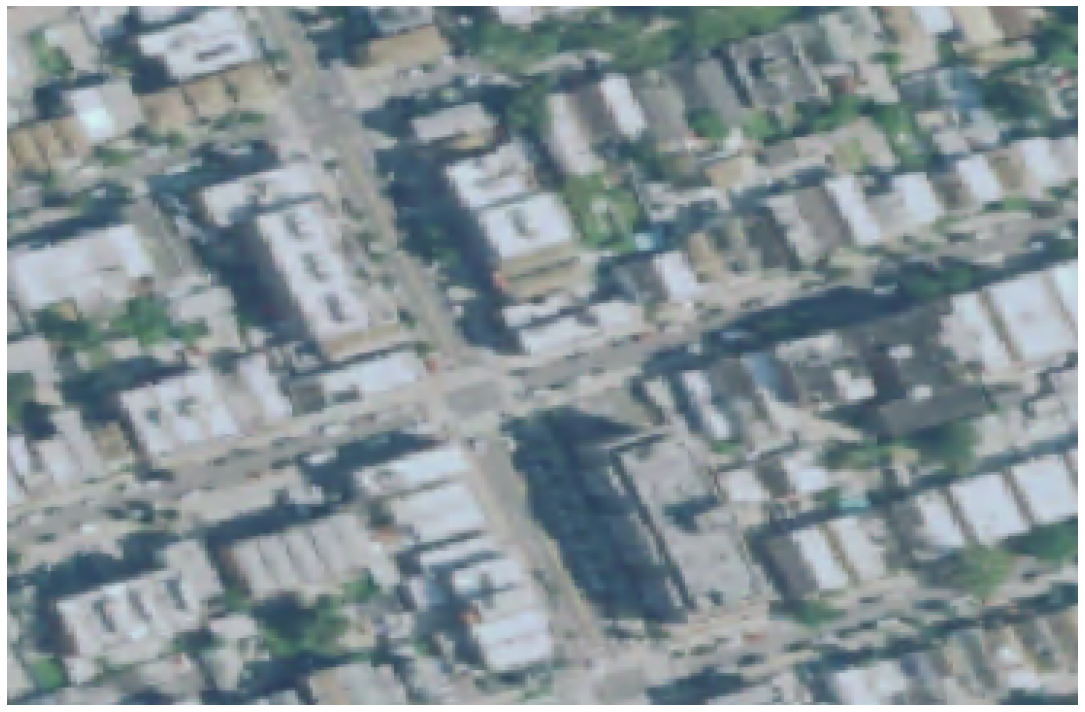}
  }
  \subfigure[rasterized LiDAR statistics]{\label{subfig:LidarStats}
      \includegraphics[width=.23\textwidth]{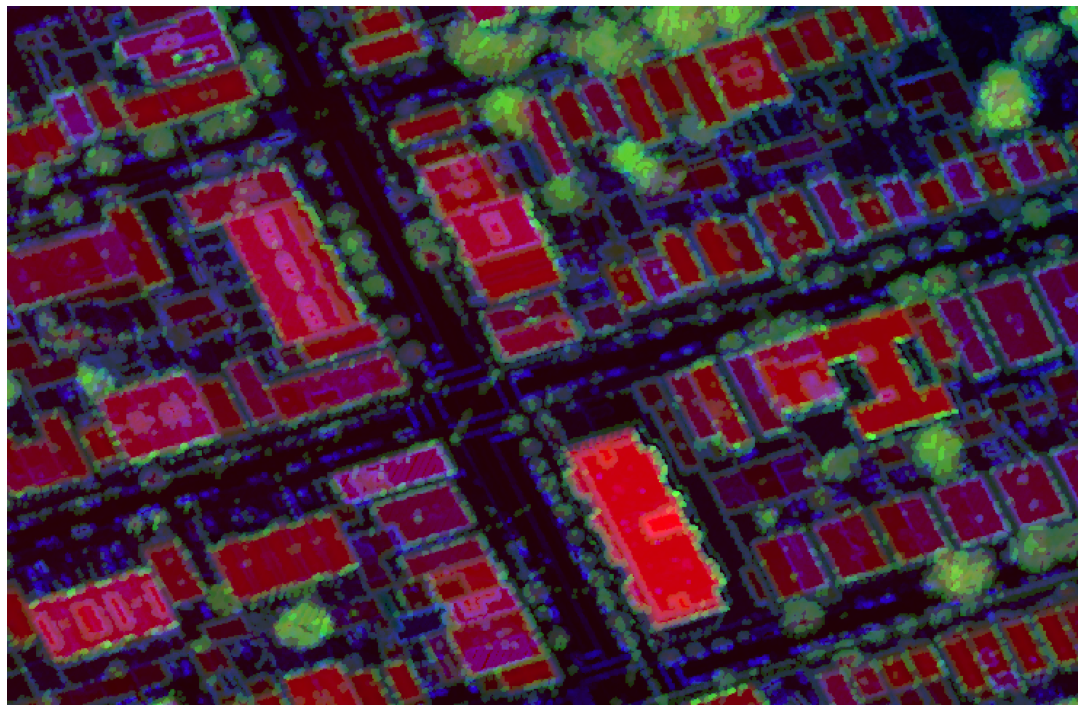}
  }
  \subfigure[LiDAR--based noisy tree labels]{\label{subfig:noisy}
     \includegraphics[width=.23\textwidth]{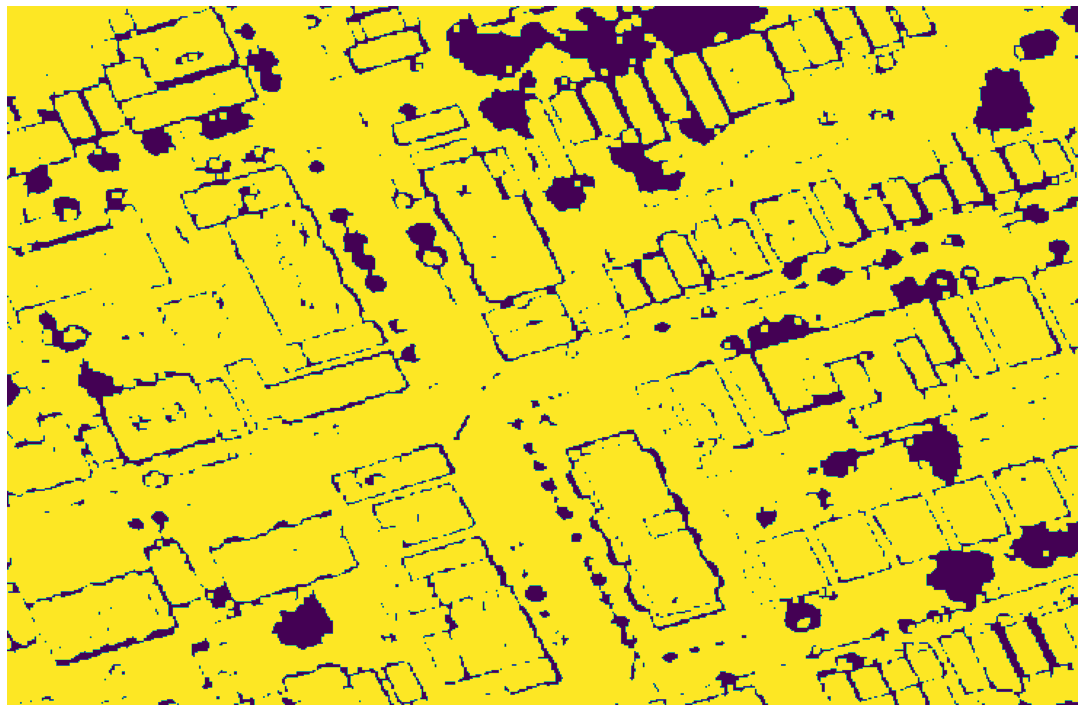}
     }
  \subfigure[exact tree labels]{\label{subfig:exact}
    \includegraphics[width=.23\textwidth]{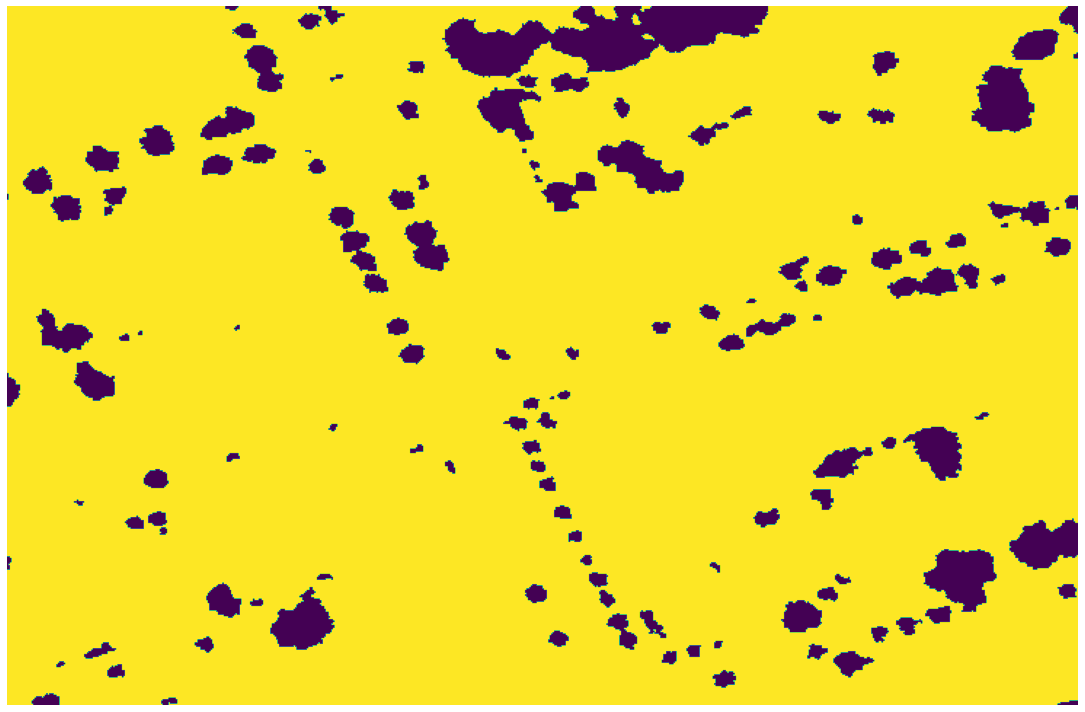}
  }
  \vspace{-2ex}
  \caption{\label{fig:DatasetAndNoisyLabels}%
  \textit{Sample patch of training data drawn from Queens borough, NYC.}
  \ref{subfig:orthophoto}: True-color RGB channels of NAIP orthophoto,
  \ref{subfig:LidarStats}: pseudo-RGB image of rasterized LiDAR statistics (Red channel: mean elevation,
  Green channel: max.\ number of laser pulse returns, Blue channel: standard deviation of laser
  reflectance), 
  \ref{subfig:noisy}: corresponding rule-based (noisy) tree segmentation map, and
  \ref{subfig:exact}: 2017 NYC Land Classification map for \textit{Tree Canopy} (dark)
  vs.\ all other classes (yellow).
  }
\end{figure*}

\subsection{New York City LiDAR survey}
Point cloud LiDAR data was acquired in 2017 \cite{nyc-lidar} at 10 points per square meter.
The raw point cloud data was converted into raster layers based on the physically sensed
laser pulse quantities, namely: \textit{time-of-flight}
(elevation), \textit{number-of-returns} (count of reflected laser pulses), and
\textit{reflected laser light intensity}. The raster layers of $0.5m$ grid size is generated
by a sliding circle with $1.5m$ diameter in order to accumulate statistics such as minimum,
maximum, mean, and standard deviation of all three physical quantities listed above,
cf. \cite{autogeolabel}. In our study, we generate noisy labels from these high-quality
LiDAR survey statistics following some physical rules detailed in \Cref{sec:methodology}.

\subsection{New York City land classification}
\label{sec:NYCLandCover}
With the aid of the 2017 New York City (NYC) LiDAR data, additional geospatial surveys like building
footprints, and overhead imagery, the city's terrain got mapped into 8 land cover classes:
\textit{Tree Canopy, Grass/Shrub, Bare Soil, Water, Buildings, Roads, Other Impervious,} and
\textit{Railroads} employing a proprietary software \cite{nyc-landcover}.
Such detailed land use information is rare to come by and costly to process. In our study, we employ
the NYC Land Classification as ground truth validation dataset for performance quantification
(cf.\ \Cref{tab:NoisyModelPerformance}) of our models trained on the noisy labels
(cf.\ \Cref{sec:methodology}).

\subsection{Multispectral imagery}
The \textit{National Agriculture Imaging Program}'s (NAIP) \cite{naip} orthophotos provide
imagery recording 4 spectral bands: near-infrared (NIR), red (R), green (G),
and blue (B). The data is acquired at $1$ meter spatial resolution every other year.
In our experiments we utilize the 2017 data for model training, and infer semantic
segmentation of trees on data of years 2011, 2013, and 2015.

Fig.\ \ref{fig:DatasetAndNoisyLabels} illustrates samples of the curated datasets for a
randomly picked geospatial area. All datasets did get spatio-temporally indexed into a
nested, common grid with the aid of open-source software libraries such as GDAL
\cite{gdal} and PDAL \cite{pdal}. Distributed data curation and model training was performed in
super-computing facilities.

\section{Methodology \& Experimental Setup}
\label{sec:methodology}

\subsection{Rule-based label generation}
After the LiDAR point cloud survey is rasterized as detailed in \Cref{sec:datasets}
with illustration in Fig.\ \ref{subfig:LidarStats},
we apply a simple, \textit{rule-based} thresholding technique to generate a (noisy) segmentation
map for trees, as depicted by Fig.\ \ref{subfig:noisy}, more details in \cite{autogeolabel}.
The rationale of this approach is guided by intuitive, physical arguments as follows:
While the laser pulse emitted by the airborne LiDAR device penetrates the tree's canopy, most
likely multiple returns get generated by partial reflections on various elevation levels
caused by branches, leaves, and potentially bare ground. In consequence, significant variation
of both, elevation and the number of returned pulses, characterizes such vegetation which,
in turn, manifests in the rasterized LiDAR statistics layers, cf.\ green channel of
Fig.\ \ref{subfig:LidarStats}.
Thresholding is sensitive to noise, generating inexact labels. As e.g.\
obvious from Fig.\ \ref{subfig:LidarStats}, edges of buildings (red channel) generate imprints in
the LiDAR statistics layers: multiple laser returns signal significant variation in elevation.
However, we note that rule-based label creation is instantaneous compared to the
lengthy process of manual labeling by visual inspections of human experts. We experimentally
quantify the trade-off \textit{noisy labeling vs.\ human inspection} in the following. 

\subsection{Experimental setup}
In a proof of concept we evaluate our approach of noisy label generation through
investigation of the damage caused by Hurricane \textit{Sandy} \cite{NYCOpenDataSandySandyInundationZone:2012},
detailed in \Cref{sec:results}.
We picked two standard machine learning models to quantify the accuracy of tree
identification from the multi-spectral NAIP imagery, Fig.\ \ref{subfig:orthophoto},
to perform supervised training on auto-generated, noisy segmentation maps, Fig.\ \ref{subfig:noisy}, namely:
(a) \textit{deep learning}: U-Net \cite{ronneberger2015u} architecture with four fully
convolutional layers of down/up sampling optimized by a mixture of standard pixel-wise cross-entropy
loss and a \textit{Dice loss} \cite{sudre2017generalised} on patches of \texttt{256$\times$256} pixels,
and
(b) \textit{standard machine learning}: Support Vector Machine (SVM) with Gaussian kernel
\cite{camps2005kernel} operating on the single-pixel, four-channel NAIP spectral space.
Once the models were successfully trained on 2017 NAIP imagery in Queens, NYC, inference
is conducted in various areas of Brooklyn, NYC for the years 2011, 2013, and 2015. We
are guided by the identification of representative locations demonstrating the effect
of Hurricane \textit{Sandy} which struck in 2012.
Spatially contrasting the inferred urban forest segmentation maps of two consecutive
temporal data (2011--2013, 2013--2015), we observe changes in tree cover as detailed in
\Cref{sec:results}, \Cref{tab:ImpactEvalSandy}, and Fig.\ \ref{fig:TreeChangeDetection}.

\section{Results \& Application}
\label{sec:results}

\subsection{Performance evaluation of noisy labels}

For identifaction of trees \Cref{tab:NoisyModelPerformance} summarizes our results
in terms of standard accuracy measures \cite{davis2006relationship} for binary
classification, specifically: precision $p$ (\textit{type I} error,
\textit{false alarm}), recall $r$ (\textit{type II} error, \textit{missed detection}), $F_1$ score ($2/F_1=1/p+1/r$), and the \textit{Intersection-over-Union} (IoU) \cite{zhang2001review}. 

\Cref{tab:NoisyModelPerformance} displays model performance based on the NYC Land Cover
dataset, cf.\ \Cref{sec:NYCLandCover} (\textit{exact}), versus the LiDAR-based
auto-generated labels (\textit{noisy}). For reference, the center row in
\Cref{tab:NoisyModelPerformance} represents performance when the noisy label generation
is considered a model itself. Three key observations we distill:
(a) Deep learning (U-Net) taking into account spatial context has an edge over
standard machine learning techniques (SVM) operating on the single-pixel level as
apparent by the IoU and $F_1$-score.
(b) Further the data support the assumption that training on \textit{exact} segmentation
maps outperforms results based on \textit{noisy} labels.
(c) However, most striking, machine learning has the potential to surpass the bare
accuracy of noisy labels trained on. Prominently, while the precision of the \textit{noisy}
labels relative to the \textit{exact} ones is at the level of 52\%, both, the SVM and
the U-Net approach generate models exceeding such by a margin of 25\% and 35\%, respectively.

\subsection{Tracing the impact of Hurricane \textit{Sandy}}
\begin{figure*}
  \centering
  \subfigure[2011: one year before storm]{
      \includegraphics[width=.23\textwidth]{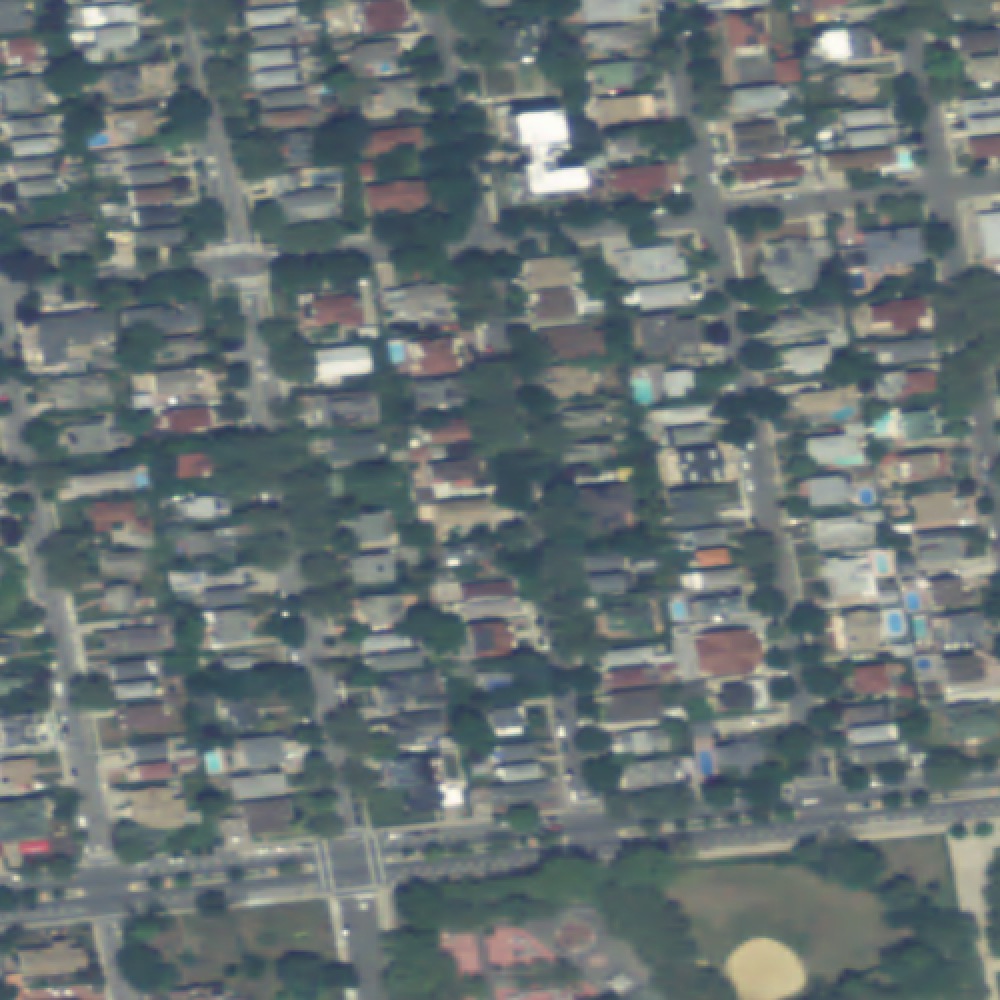}
  }
  \subfigure[2013: one year after storm]{
      \includegraphics[width=.23\textwidth]{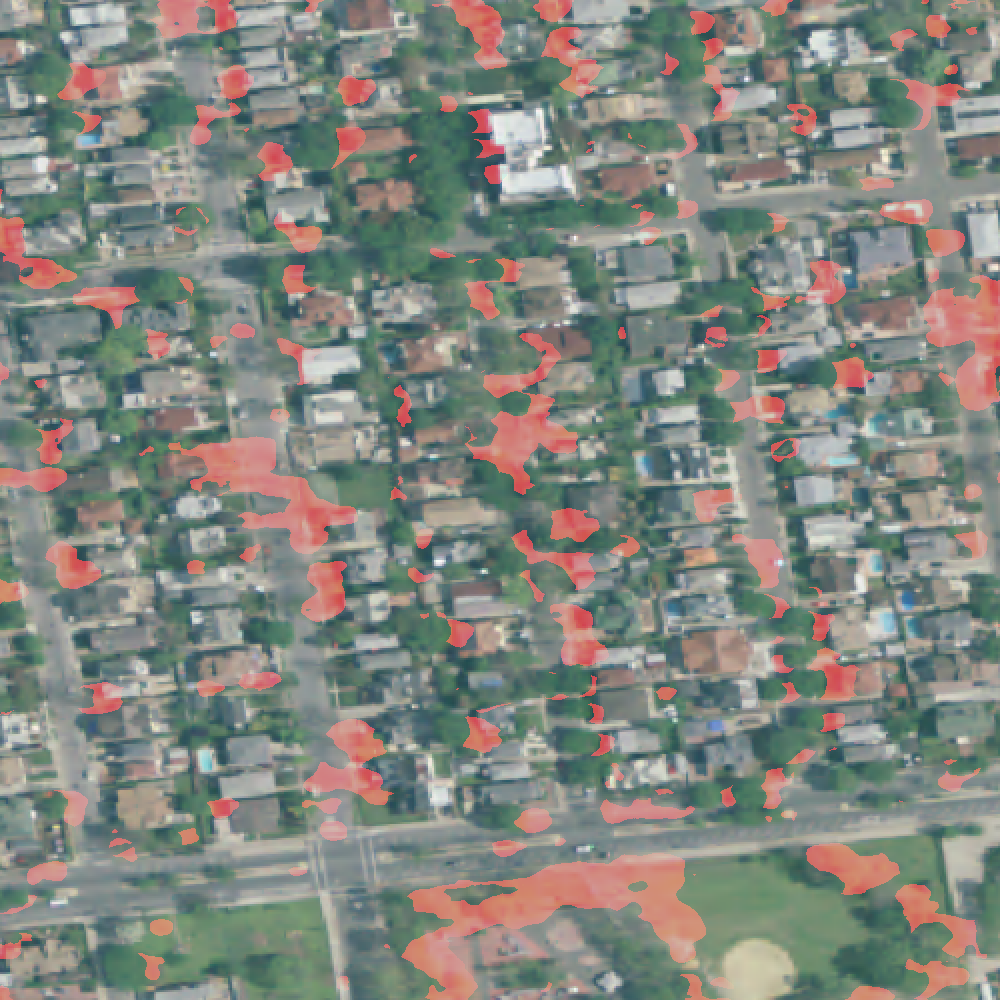}
  }
  \subfigure[2015: 3 years after storm]{
      \includegraphics[width=.23\textwidth]{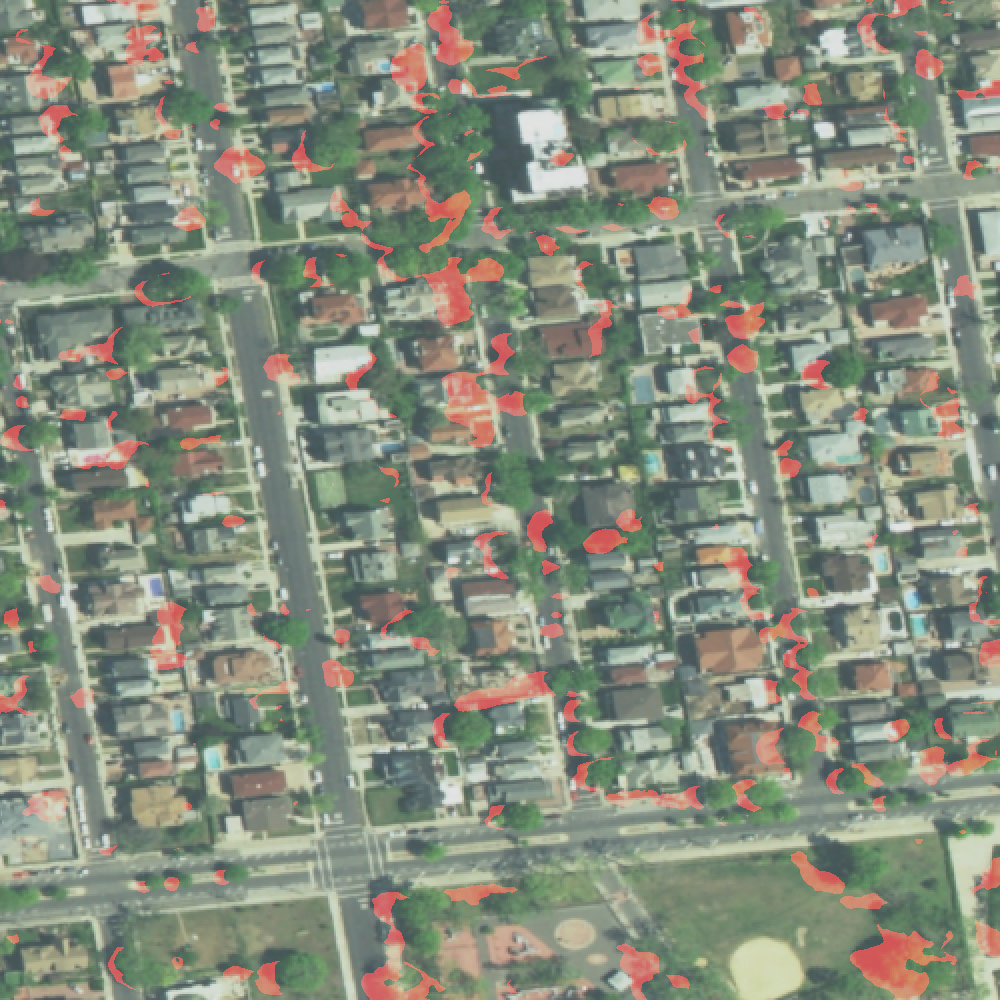}
  }\\[.5ex]
  \vspace{-2ex}
  \caption{\label{fig:TreeChangeDetection}%
  \textit{Visualization of Hurricane Sandy's impact on Coney Island and inland Brooklyn, NYC.}
  Columns encode years 2011, 2013, and 2015 from left to right. It is depicted
  the RGB channels of NAIP orthophotos with semi-transparent, red overlay of fallen
  trees as contrasted by its neighboring picture to the left, i.e. the previous
  temporal snapshot.}
\end{figure*}

In order to investigate Hurricane \textit{Sandy}'s damage to the urban forests of
NYC, we let the U-Net model detect trees in Brooklyn before and after the extreme
weather event. Through such survey we identified 3 qualitatively separate scenes
as summarized in \Cref{tab:ImpactEvalSandy}: (a) suburban areas close by the
sea side with a significant loss in forests, (b) urban areas with less total area
of tree cover, but protective blocks of massive, high-elevation buildings
that seem to mitigate damage in trees, (c) suburban neighborhoods inland with
little to no loss in urban forests.
In fact, tracking the change in urban tree cover over consecutive years may reveal
secondary effects. Specifically, while the change from 2011 to 2013 indicates
direct damage by the storm, the 5\% loss observed from 2013 to 2015 is likely from
storm-caused flooding near the shore where salty sea water lets malfunction roots of
trees long-term. 
To illustrate, Fig.\ \ref{fig:TreeChangeDetection} provides several snapshots of a
highly impacted suburban area before and after the storm. As observed, our model is
mature to generate reasonable detection results of fallen trees after Hurricane \textit{Sandy}.

\section{Conclusion \& Perspectives}

Rapid generation of geospatial labels is essential to analyze overhead imagery for change detection. It bears potential for environmental analysis of our planet, to e.g.\ provide clues on climate--induced land surface change: icecap melting, greening of the arctic, and patterns of vegetation type change due to raising temperatures and drought---to name a few. Here we presented a method to quickly generate \textit{noisy} labels from one data type, and use such labels for another data type with little human interaction to delineate urban forests. Semantic segmentation using U-Net and SVM learning indicate comparable results to models trained on human expert–labeled data. In a proof-of-concept study, the change due to 2012 Hurricane \textit{Sandy} was quantified for the New York City coastal regions.

\begin{table}[t!]
  \caption{\label{tab:NoisyModelPerformance}%
  Model performance measures for tree identification.}
  \centering
  \scalebox{.9}{
    \begin{tabular}{|c|c||c|c|c|c|}
      \hline
      Model & Label & precision & recall & $F_1$ & IoU \\\hline
      \multirow{2}{*}{\bf U-Net}
        &\it exact & .85 & .81 & .82 & .71 \\\cline{2-6}
        &\it noisy & .87 & .60 & .69 & .55 \\\hline\hline
      \multicolumn{2}{|c||}{ref.\ \textit{noisy} vs.\ \textit{exact}\cite{autogeolabel}}
                   & .52 & .60 & .55 & .38 \\\hline\hline
      \multirow{2}{*}{\bf SVM}
        &\it noisy & .77 & .58 & .63 & .48 \\\cline{2-6}
        &\it exact & .82 & .54 & .62 & .48 \\\hline
      \end{tabular}
  }
\end{table}

\begin{table*}[t!]
  \caption{\label{tab:ImpactEvalSandy}%
  Impact of Hurricane \textit{Sandy} (2012) on trees in Brooklyn neighborhoods, NYC.}
  \centering
  \small
    \begin{tabular}{lccrr}
      \hline
      \bf scene definition              &
      \bf sample reference              &
      \bf vegetation state              &
      \multicolumn{2}{c}{\bf relative change}\\
                                        &
      $(lat, lon)$                      &
      \bf description                   &
      $2011\rightarrow2013$ & $2013\rightarrow2015$\\
      \hline\hline
      suburban close to shore & \href{https://www.openstreetmap.org/#map=18/40.57896/-73.94853}{$(40.579,-73.947)$} &
      \it damaged          & -14\%  & -5\%\\
      urban close to shore              & \href{https://www.openstreetmap.org/#map=18/40.57936/-73.95867}{$(40.580,-73.958)$} &
      \it little damage    & -3\%   & +4\%\\
      suburban inland                   & \href{https://www.openstreetmap.org/#map=18/40.59865/-73.96845}{$(40.599,-73.967)$} &
      \it undamaged        &  +.5\% & +3\%\\
      \hline
    \end{tabular}\\
  \footnotesize
  results based on U-Net inference on NAIP imagery of years 2011, 2013, and 2015, geolocation hyperlinked to OpenStreetMap web page
  \vspace{-2ex}
\end{table*}

Our initial experiments call for future systematic benchmarks of the approach: How may we adapt machine/deep learning architectures to cope with the noisy nature of labels to improve model accuracy for change detection in urban forests? To which extent may self-supervised learning \cite{jing2020self} assist in mitigating inference noise from orthophotos with variation in e.g. seasonal conditions of the tree's canopy? Moreover, regarding the analysis of damage caused by \textit{Sandy}: We have a mild indication that trees along roads from South--North is more severely damaged compared to their East--West counterparts in the Eastern Coney Island suburbs. 
Urban planning may benefit from automated remote sensing to identify and mitigate the impact of natural disasters, and to efficiently drive reforestation.
Since NAIP data and LiDAR surveys is available for major US cities, future work
may apply the proposed method to various geospatial locations affected by natural
hazards in order to proof the efficiency of our approach.

{
\small
\bibliography{ref.bib}
\bibliographystyle{unsrt}
}

\end{document}